\documentclass[journal]{IEEEtran}
\usepackage[T1]{fontenc}
\usepackage{mathptmx}
\usepackage{amsmath,amsfonts}
\usepackage{algorithmic}
\usepackage{algorithm}
\usepackage{array}
\usepackage[caption=false,font=footnotesize,labelfont=rm,textfont=rm]{subfig}
\usepackage[justification=centering]{caption}
\usepackage{textcomp}
\usepackage{multirow}
\usepackage{stfloats}
\usepackage{url}
\usepackage{verbatim}
\usepackage{graphicx}
\usepackage{booktabs}
\usepackage{amssymb}
\usepackage[style=ieee, doi=false, url=false]{biblatex}  
\begin{document}
\title{A cross-modal network for facial expression recognition}

\author{
    Chunwei Tian, ~\IEEEmembership{Senior Member,~IEEE,} Jingyuan Xie, Qi Zhang, Chao Li, \\
    Wangmeng Zuo, ~\IEEEmembership{Senior Member,~IEEE,} and Shichao Zhang ~\IEEEmembership{Senior Member,~IEEE} 
    \thanks{
        This work was supported in part by National Natural Science Foundation of China under Grant 62576123, and part by Natural Science Foundation of Heilongjiang Province under Grant YQ2025F003. (Corresponding author: Qi Zhang.)
        
    }
    \thanks{
        Chunwei Tian is with the School of Computer Science and Technology, Harbin Institute of Technology, Harbin 150001, China and the School of Software, Northwestern Polytechnical University, Xi'an 710072, China (e-mail: chunweitian@hit.edu.cn).
    }
    \thanks{
        Jingyuan Xie is with the School of Software, Northwestern Polytechnical University, Xi'an 710072, China (e-mail: xjyjason2001@mail.nwpu.edu.cn).
    }
    \thanks{
        Qi Zhang is with the School of Economics and Management, Harbin Institute of Technology at Weihai, Weihai, 264209, China (e-mail: hit\_zq910057@163.com).
    }
    \thanks{
        Chao Li is with the School of Computer Science and Engineering, Central South University, Changsha, China (e-mail: 1372116696@qq.com).
    }
    \thanks{
        Wangmeng Zuo is with the School of Computer Science and Technology, Harbin Institute of Technology, Harbin 150001, China (e-mail: cswmzuo@gmail.com).
    }
    \thanks{
        Shichao Zhang is with the Key Lab of MIMS, College of Computer Science and Technology, Guangxi Normal University, Guilin 541004, China (e-mail: zhangsc@mailbox.gxnu.edu.cn).
    }
}

\maketitle

\begin{abstract}
    Deep neural networks enriched with structural information have been widely employed for facial expression recognition tasks. However, these methods often depend on hierarchical information rather than face property to finish expression recognition. In this paper, we propose a cross-modal network with strong biological and structural information for facial expression recognition (CMNet). CMNet can respectively learn expression information via face symmetry on a whole face, left and right half faces to extract complementary facial features. To prevent negative effect of biological and structural information fusion, a salient facial information refinement module can obtain salient facial expression information to improve stability of an obtained facial expression classifier. To reduce reliance on unilateral facial features, a half-face alignment optimization mechanism is designed to align obtained expression information of learned left and right half faces. Our experimental results demonstrate that CMNet outperforms several novel methods, i.e., SCN and LAENet-SA for facial expression recognition. Codes can be obtained at https://github.com/hellloxiaotian/CMNet.
\end{abstract}


\section{Introduction}
\IEEEPARstart{F}{acial} expressions directly reflect people's current emotions to easily communicate with different people \cite{shatzman_expression_1993}. With the development of internet technique, human-computer interaction progressively becomes a mainstream communication way rather than face-to-face. Given its potential to enhance social communication and enable wide applications, understanding emotional data is of great importance \cite{zhao_2016_predicting}, \cite{zhao_2023_toward}. This understanding can enhance social communication and enable a wide range of applications, which necessitates robust methods for emotion analysis, positioning facial expression techniques as a fundamental tool. Ekman et al.  \cite{ekman_constants_1971} established a connection between facial expressions and emotions in their pioneering work and categorized them into six fundamental types: anger, disgust, fear, happiness, sadness, and surprise. Subsequently, neutral and contempt expressions are also added to conduct public facial expression recognition (FER) datasets to train a classifier to classify facial images or video clips to automatically determine their fundamental expression categories. Taking highly similar characteristics of different facial images and a common underlying facial appearance feature in all the expression categories into account, distinction between expression categories lied in the combination of various action units are conducted to recognition facial expressions. Deep learning with deep architectures can effectively learn subtle variations of facial muscle movements from data dependency for facial expression recognition.

Attention mechanisms implemented by perception ideas can extract salient information to highlight facial features for FER \cite{guo_attention_2022}. Wang et al. presented a coarse-to-fine mechanism to establish a relation between different image blocks to address covered facial expression recognition \cite{wang_region_2020}.  That is, the first step utilized an attention method to act randomly cropping on image blocks, mining their relations to obtain local salient facial expression features to overcome loss information of occlusion in FER. The second step aggregated these obtained local features to construct the whole feature image for improving robustness of FER.  To improve efficiency of FER, reducing network depth can simplify an attention mechanism to efficiently mine emotionally significant areas within images for FER \cite{wang_light_2021}. Also, a cross-entropy can assist the mentioned operation to address unbalanced samples caused by non-standard collection data to improve effectiveness of FER. Alternatively, Farzaneh et al. \cite{farzaneh_facial_2021} introduced a sparse center loss via considering the contribution of each dimension in deep expression feature vectors obtained by a CNN to generate highly discriminative expression features for FER. Additionally, integrating auxiliary information branches, i.e., manual patch selection \cite{zhao_learning_2021}, automatic acquisition based on facial landmarks \cite{li_occlusion_2018} can guide a CNN to focus on specific regions for learning more facial expression information in FER. Although these methods can improve recognition rate of FER, they usually required manual setting parameters, i.e., the size and number of selected regions within region-based attention mechanism , which may limit their generalization ability across different scenarios.

To tackle issue of high intra-class similarity and low inter-class discrepancy in facial expression recognition, we use two modals obtained by hierarchical relations in a deep network and face property to construct an efficient model for FER. That is, we present a cross-modal network (CMNet) with biological and structural information for FER. According to face symmetry, CMNet uses a multi-view idea to guide a CNN to respectively learn more facial expression information from the whole face together with left and right half faces to facilitate complementary facial features for FER. To prevent negative effect of biological and structural information fusion, a salient facial information refinement module is designed to mine salient facial expression information to improve stability of an obtained facial expression classifier. To prevent asymmetry of cropped half face images, a half-face alignment optimization mechanism is used to align obtained facial expression information from learned left and right half features. Experimental results show that our CMNet is superior to several popular methods, i.e., SCN and LAENet-SA for facial expression recognition.

This paper has the following contributions.

1) A cross-modal technique can facilitate an interaction between biological and structural information to achieve a stable model for facial expression recognition.

2) To prevent negative effect of cross-modal information, a salient facial information refinement module is used to further extract representative facial expression information to ensure performance of facial expression classifier.

3) To prevent reliance on global facial information, half-face alignment optimization mechanism is proposed to align obtained facial information to improve accuracy of facial expression recognition.

Remaining parts of this paper can be shown as follows. Section II refers to related work of facial expression recognition. Section III gives more detailed information of the proposed CMNet. Section IV illustrates datasets, experimental settings, the proposed method analysis and performance of our CMNet in facial expression recognition. Section V concludes the whole paper.

\section{Related Work}
\subsection{Traditional machine learning for facial expression recognition}
Early research on facial expression recognition primarily relied on handcrafted features and traditional machine learning classifiers, extracting geometric, texture, or appearance information from static images to distinguish between different expressions \cite{zhao_2021_affective}. These methods made full use of prior knowledge about the face, aiming to capture discriminative features associated with emotional expression. In terms of texture feature extraction, Gabor wavelets were widely employed to extract multi-scale and multi-directional texture information owing to their excellent selectivity for orientation and frequency to capture subtle facial expression changes \cite{lyons_coding_1998}, \cite{buciu_2003_ica}. Lyons et al. \cite{lyons_coding_1998} firstly introduced Gabor wavelets into the field of facial expression coding and achieved effective facial expression classification results by extracting Gabor magnitude features from facial key points. Building on this foundation, Buciu et al. \cite{buciu_2003_ica} compared the Gabor transform with independent component analysis, validating its effectiveness in FER tasks. To address complex conditions such as illumination variations and occlusions, sparse representation theory has been introduced into the field of FER. Cotter \cite{cotter_2010_sparse} extended the sparse representation classification method to FER tasks under occlusion and distortion, effectively improving model performance in complex conditions by incorporating robust processing of occluded regions during dictionary construction. Ouyang et al. \cite{ouyang_2015_accurate} further proposed a sparse representation method based on multi-classifier fusion, which significantly enhanced recognition accuracy and robustness to noise by fusing the decision outcomes of multiple sparse representation classifiers and leveraging complementary information from different dictionaries and feature spaces. To address the challenges posed by individual differences in personalized expression analysis, Chu et al. \cite{chu_2016_selective} constructed personalized classifiers and performed weighted transfer using samples from the source domain that are highly similar to the target domain to effectively resolve the distribution inconsistency between training data and test subjects, significantly improving the accuracy of cross-subject expression recognition. To further explore the underlying characteristics of expression variations, methods based on manifold learning have been introduced into FER tasks to enhance classification performance. Zheng et al. \cite{zheng_2015_two} proposed a two-dimensional discriminant multi-manifold locality preserving projection method, which constructed an adjacency graph to model the diversity and similarity of salient regions within the same expression class, effectively improving facial expression recognition performance. Multi-feature fusion strategies have also been widely adopted. Barman and Dutta \cite{barman_2021_facial} combined distance features with shape signature features and achieved effective expression classification using a multi-class support vector machine. Although traditional machine learning methods rely on handcrafted features and perform optimally only in controlled environments, they establish crucial prior knowledge about the importance of facial regions for expression analysis \cite{sajjad_comprehensive_2023}. Methods such as Gabor wavelet extraction from facial key points \cite{lyons_coding_1998} and sparse representation with occlusion handling \cite{cotter_2010_sparse} demonstrate that local facial regions carry discriminative emotional information and that multi-view strategies can enhance robustness. Inspired by these foundational insights, our CMNet directly integrates several key ideas from traditional machine learning. By embedding these traditional principles within a modern deep learning framework, our approach not only pays tribute to prior knowledge but also significantly improves the ability to capture salient facial expression features, effectively bridging classical wisdom with contemporary architectures for superior FER performance in real-world scenarios.

\subsection{Deep networks for facial expression recognition}
With the continuous expansion of data, deep learning techniques with automated feature extraction and data-driven autonomous learning have witnessed increasingly widespread application in FER task. Specifically, end-to-end architectures like CNNs can effectively extract structural and textual information from images to better express human faces in FER. These methods based on CNNs can be divided into two kinds, i.e., face information enhancement and salient emotion extraction for complex FER task, i.e., occlusions and varying poses.

Face information enhancement techniques refer to multi-source information to obtain more complementary detailed information for improving robustness of FER classifiers. For instance, Xie et al. \cite{xie_adaptive_2019} depended on a two-phase frame containing handcrafted feature extraction and automatic feature learning to obtain an accurate FER method. That is, the first phase utilized Gabor, GSF and SIFT to extract facial prior information. The second phase employed obtained facial prior information as the input to restrict a CNN to better learn more representative facial information in FER. Wu et al. \cite{wu_-net_2023} utilized a landmark space to extract facial key point information to enhance the whole face image for improving emotion recognition ability. Recently, researchers  proposed a method using five parallel networks to extract local features from organs such as eyes, cheeks, and mouth \cite{tian_2025_perception}. These local features are then registered and fused with global facial structural features via a multi-domain interaction mechanism, enabling sensitive capture of subtle expression variations. To obtain more recognition information, Ni et al. \cite{ni_facial_2022} fused multi-modal information containing RGB gray images, local binary pattern images, and depth images via multi-path networks to address facial expression recognition in challenging conditions, i.e., varying illuminations, diverse poses and different scenarios. Alternatively, Ruan et al. \cite{ruan_feature_2021} exploited isomorphism and heterogeneity of different facial expressions to facilitate more complementary facial detailed information composed of facial movement and structural information  for FER. To resolve occluded facial image recognition problem, binary masks were utilized to fill lack areas to strengthen emotional information for improving robustness of FER \cite{huang_fermixnet_2024}.

Salient emotion extraction techniques exploit relations of structural information to extract key facial information to guide CNNs for promoting performance of FER algorithms. Specifically, CNNs can establish models between salient emotion information and facial images in FER via using an attention mechanism in general. Li et al. \cite{li_learning_2021} designed a multi-path attention CNN with a balanced separation loss to simultaneously optimize intra-class compactness and inter-class separability to better achieve interactions between salient network width information for enhancing performance of FER in real-world scenarios. Similarly, Ye et al. \cite{ye_2025_cmdvit} proposed explicit visual center position encoding combined with an implicit sparse attention center loss to alleviate inter-class similarity and intra-class variance issues in FER, further enhancing model robustness through refined feature learning and loss optimization. To reinforce relations of multi-domain structural information, Zhao et al. \cite{zhao_learning_2021} applied spatial and channel attention mechanisms to optimize a CNN to focus on facial information in improving performance of FER. Yu et al. \cite{yu_co-attentive_2022} designed a soft parameter sharing mechanism based on a hybrid attention to unify two tasks, i.e., FER and facial landmark detection into a framework to reduce effects of pose variations and occlusion. To suppress effects of irrelevant regions, Gao et al. \cite{gao_jadfer_2024} utilized discarding activation to optimize an attention mechanism to reduce negative influence of background for addressing FER issue. In addition, Wang et al. \cite{wang_2026_mhan} incorporated an efficient local attention hybrid feature network and a multi-head hybrid attention mechanism, further boosting the model's facial expression detail capture capability. To decrease complexity of obtained FER algorithms, modeling local and global information can be presented to make a trade-off between obtained salient information and complexity of FER methods \cite{chen_multi-relations_2023}. Xia et al. \cite{xia_relation-aware_2021} fused chosen key local structural features via a fixed cropping strategy and MLP to enhance a relation between key regions and expressions in FER. Alternatively, Liu et al. \cite{liu_adaptive_2022} can gather local information obtained by a facial landmark methods and global information extracted by a multi-layer network to quickly establish a model between key regions and salient emotions for FER. Due to small sample issue of facial expression recognition, face information enhancement techniques above are essential to enrich detailed information to better express face images for FER. Also, facial organs can assist FER algorithms to easily distinguish different expressions. Thus, salient emotion extraction techniques above are useful for FER. Inspired by that, we achieve a cross-modal technique via interacting biological and structural information to obtain more detail information and improve an attention mechanism to mine more salient emotional information for better recognizing facial expressions in this paper.
\begin{figure*}[!t]
    \centering
    \includegraphics[width=\textwidth]{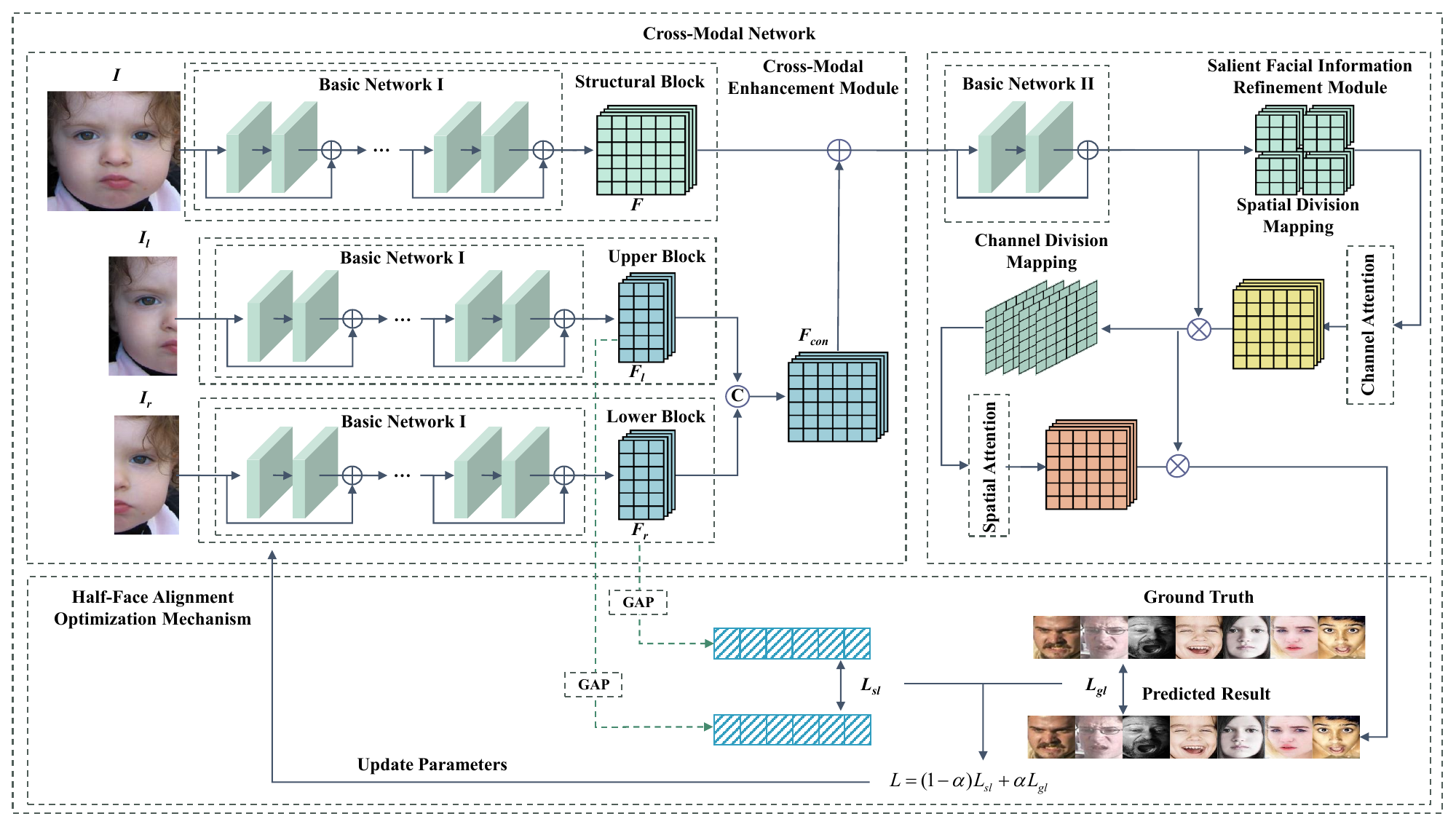}
    \caption{The architecture of the proposed CMNet for facial expression recognition.}
    \label{structure}
\end{figure*}
\section{Proposed method}
\subsection{Network architecture}
The proposed CMNet is composed of a cross-modal enhancement module (CMEM), a salient facial information refinement module (SFIRM) and a half-face alignment optimization mechanism (HFAOM) in Fig. \ref{structure}. To improve robustness of an obtained facial expression recognition model, CMEM is presented via fusing structural and biological information to extract complementary emotional information for facial expression recognition. That is, structural information can be obtained via a structural block (SB) based on residual networks \cite{he_deep_2016} to extract common facial information for FER. Also, biological information can be obtained via using parallel biological blocks based on the same residual networks above \cite{he_deep_2016} regarded as upper and lower blocks (UB and LB) to process left and right half face images to learn symmetric facial information for FER. The two-part generalized modal information is more beneficial to express the whole face image for FER. Finally, we use a two-phase fusion mechanism to fuse two modal information to enhance facial information. The first phase uses a concatenation operation to splice obtained information from the UB and LB to guarantee the size same as that of the SB. The second phase utilizes a residual operation to gather obtained information from the first phase and SB. More detailed information of CMEM can be shown in Section III. B. To prevent over-enhancement of cross-modal enhancement module, a salient facial information refinement module is proposed. SFIRM applies a cross mapping to extract cross-domain information for refining salient emotion facial information in FER, which can be illustrated in Section III. D. Besides, to improve symmetry of left and right half faces, a half-face alignment optimization mechanism (HFAOM) via a perception loss is proposed. That is, HFAOM utilizes differences between obtained facial features from left and right half faces through UB and LB to restrict symmetry of half face images for FER. These features are first subjected to Global Average Pooling (GAP) processing, and then subjected to symmetric perceptual loss constraints, updating model parameters together with cross entropy loss, as shown in Fig. \ref{structure}. The specific description can be found in Section III.C. Mentioned descriptions can be visualized as the following equation.
\begin{equation}
    \begin{aligned}
        O & = CMNet(I, I_l, I_r)                                     \\
          & = SFIRM(CMEM(I, I_l, I_r)) \& HFAOM(O_{SFIRM}, F_l, F_r) \\
          & = SFIRM(O_{CMEM})          \& HFAOM(O_{SFIRM}, F_l, F_r) \\
          & = SFIRM(O_{CMEM})          \& O_{HFAOM}                  \\
    \end{aligned}
\end{equation}

where  $CMNet, CMEM, SFIRM, HFAOM$ denote functions of CMNet, SFIRM, CMEM, SFIRM and HFAOM, respectively.  $I, I_l, I_r$ are used to express a whole face image, a left half face image and a right half image, respectively. $O, O_{SFIRM}$ are outputs of CMNet and SFIRM, respectively. $O_{CMEM}$ is an output of CMEM. $O_{HFAOM}$ is an output of HFAOM. $F_l$ and $F_r$ represent feature mapping extracted from left and right faces through UB and LB, respectively. $\&$ stands for optimizing parameters of CMNet.

\subsection{Cross-modal enhancement mechanism}
To improve accuracy of FER, the cross-modal enhancement mechanism uses an interaction between structural and biological information to capture more emotional information for FER. Structural information is mined by a structural block based on a 14-layer residual network (also regarded as Basic Network I) \cite{he_deep_2016}. Biological information $F$ can be captured by parallel blocks based Basic Network I, i.e., the Upper Block (UB) and Lower Block (LB). That is, UB and LB can learn symmetric facial information for FER. To keep the same size as the SB, we utilize a concatenation operation to splice obtained information from a UB and LB, i.e., $F_l$ and $F_r$ to obtain new features as $F_{con}$. To fuse two different modal information, a residual learning operation is used to act $F$ and $F_{con}$ to obtain a cross-modal enhancement information. Mentioned illustrations can be equaled as Eq. \ref{eq2}.
\begin{equation}
    \begin{aligned}
        O_{CMEM} & = CMEM(I, I_l, I_r)                \\
                 & = SB(I) + Concat(UB(I_l), LB(I_r)) \\
                 & = F + Concat(F_l, F_r)             \\
                 & = F + F_{con}                      \\
    \end{aligned}
    \label{eq2}
\end{equation}
where $SB$, $UB$ and $LB$ represent functions of SB, UB and LB, respectively. $Concat$ is a concatenation operation. $+$ is a residual learning operation. Basic Network I is a 14-layer residual network, where its more detailed information can be shown in Ref. \cite{he_deep_2016}.

\begin{figure}[htbp]
    \centering
    \includegraphics[width=\linewidth]{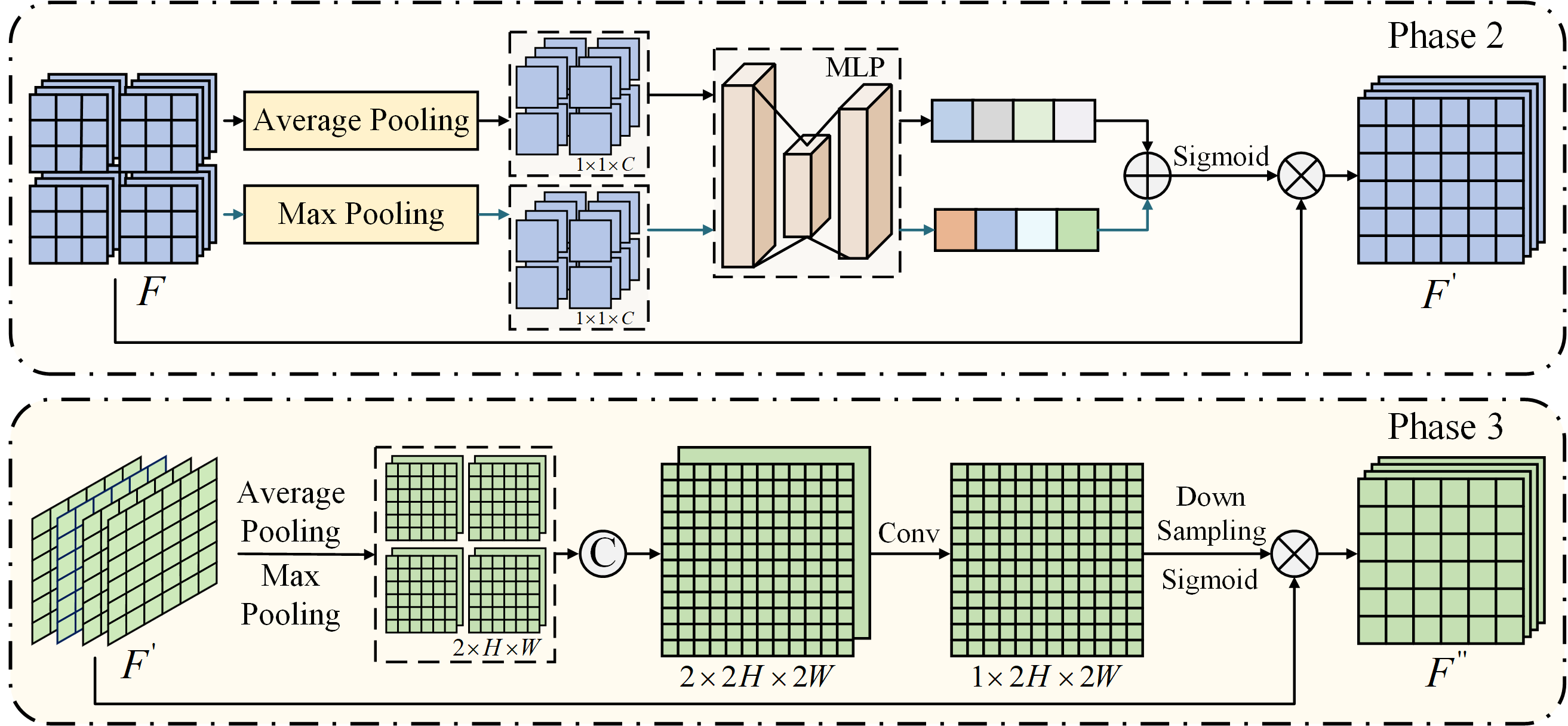}
    \caption{The architecture of the proposed SFIRM for facial expression recognition without Basic Network II.}
    \label{SFIRM}
\end{figure}
\subsection{Salient facial information refinement mechanism}
To remove redundant information from cross-modal enhancement module, the salient facial information refinement module (SFIRM) is presented. SFIRM exploits different domain information interactions via three phases to extract salient facial information for FER.  
The first phase utilizes 4-layer residual network named Basic Network II to refine obtained information from the CMEM, where more information of Basic Network II can be obtained in Ref. \cite{he_deep_2016}. The second phase utilizes interactions between local spatial features and global channel features, where its visual figure is given in Fig. \ref{SFIRM}. That is, firstly, a division method of central point can be used to split spatial features to four parts. Secondly, it uses a channel attention mechanism \cite{woo_cbam_2018} to extract four local salient channel information for FER. After that, to keep same size as the input image, obtained four features can be joined as a whole feature. Finally, we use a common attention mechanism to interact refined information from Basic Network II and obtained whole feature from the last operation. The third phase depends on interactions between local channel features and global spatial features. That is, firstly, obtained features from last operation can be divided into four channel mappings. Secondly, four obtained channel mappings can learn salient spatial information by a spatial attention \cite{woo_cbam_2018}. Thirdly, to keep same size as an input image, four learned spatial mappings can be joined as a whole feature mapping, where its visual figure is given in Fig. \ref{SFIRM}. The process mentioned can be shown as follows.
\begin{equation}
    \begin{aligned}
        O_{SFIRM} & = SFIRM(O_{CMEM})                                        \\
                  & = M(J(SA(CDM(M(J(CA(SDM(BN_{II}(                         \\
                  & \quad \quad O_{CMEM})))), BN_{II}(O_{CMEM}))))), O_{se}) \\
                  & = M(J(SA(CDM(O_{se}))), O_{se})                          \\
    \end{aligned}
\end{equation}
where $BN_{II}$, $SDM$, $CA$, $J$, $M$, $CDM$ and $SA$ stand for functions of Basic Network II, spatial division mapping, channel attention, joint, multiplication, channel division mapping and spatial attention, respectively. $O_{se}$ is an output of the second phase in the SFIRM. $O_{SFIRM}$ is final output of SFIRM.

\subsection{Half-face alignment optimization mechanism}
To encourage CMNet to effectively learn facial symmetry information of left and right half faces, a half-face alignment optimization mechanism (HFAOM) is proposed. HFAOM consists of two parts: a symmetry loss and a global loss. The symmetry loss is utilized to restrict symmetry of half face images for FER. That is, it uses global average pooling to deal with obtained features from the UB and LB to refine salient information. Then, it utilizes a variance function to measure difference between obtained left and right half face features to guarantee symmetry of half faces for FER. The global loss is utilized to evaluate effect of proposed CMNet for facial recognition. It mainly depends on differences between outputs of SFIRM and given ground truth to verify performance of obtained CMNet in FER. Mentioned descriptions can be visualized as the following equation.
\begin{equation}
    \begin{aligned}
        O_{HFAOM} & = HFAOM(O_{SFIRM}, F_l, F_r)        \\
                  & = (1-\alpha) L_{sl} + \alpha L_{gl}
    \end{aligned}
    \label{eq4}
\end{equation}
where $L_{sl}$ and $L_{gl}$ stand for a symmetry loss and global loss, respectively. $\alpha$ is a weighting hyper-parameter, which is exploited to balance two losses. Also, $\alpha$ is set to 0.9 in this paper. $L_{sl}$ depends on difference of symmetrical feature pixel points to restrict symmetry of left and right half faces.

It is implemented by the following three parts. First part exploits average pooling operation to decrease and refine obtained symmetrical half-face features. Detailed equation can be listed as follows.
\begin{equation}
    \begin{cases}
        v_l = GAP(F_l) \\
        v_r = GAP(F_r)
    \end{cases}
\end{equation}
where $GAP$ is a function of global average pooling operation. $v_l$ and $v_r$ are vectors of decreased symmetrical features, respectively. Second part uses an adjusted activate function based on Softmax \cite{bridle_probabilistic_1990} to normal obtained symmetrical features as Eq. \ref{eq6}.
\begin{equation}
    \begin{cases}
        x_l = \log\left(\frac{e^{v_l}}{e^{v_l} + e^{v_r}}\right) \\
        x_r = \log\left(\frac{e^{v_r}}{e^{v_l} + e^{v_r}}\right)
    \end{cases}
    \label{eq6}
\end{equation}
where $x_l$ and $x_r$ are corresponding feature vectors of left and right half-faces, respectively. Third part uses Eq. \ref{eq7} to act corresponding feature pixel points to measure difference of obtained symmetrical half-face features.
\begin{equation}
    L_{sl} = \frac{2}{NC} \sum_{i=1}^{N} \sum_{j=1}^{C} (x_l^{(i,j)} - x_r^{(i,j)})^2
    \label{eq7}
\end{equation}
where $x_l^{(i,j)}$ and $x_r^{(i,j)}$ denote pixel points of $(i, j)$ in the left and right half-faces, respectively. $C$ represents channel number of obtained $x_l$ and $x_r$. And $N$ denotes batch size of samples. The global loss is based on the cross-entropy loss, whose more detailed information can be found in Ref. \cite{lee_context-aware_2019}.

\section{Experimental analysis and results}
\subsection{Datasets}
To fully test performance of our CMNet for FER, we choose different scenes, i.e., real scenes and context-sensitive scenes to conduct experiments in this paper.

\begin{figure}[h]
    \centering
    \includegraphics[width=\linewidth]{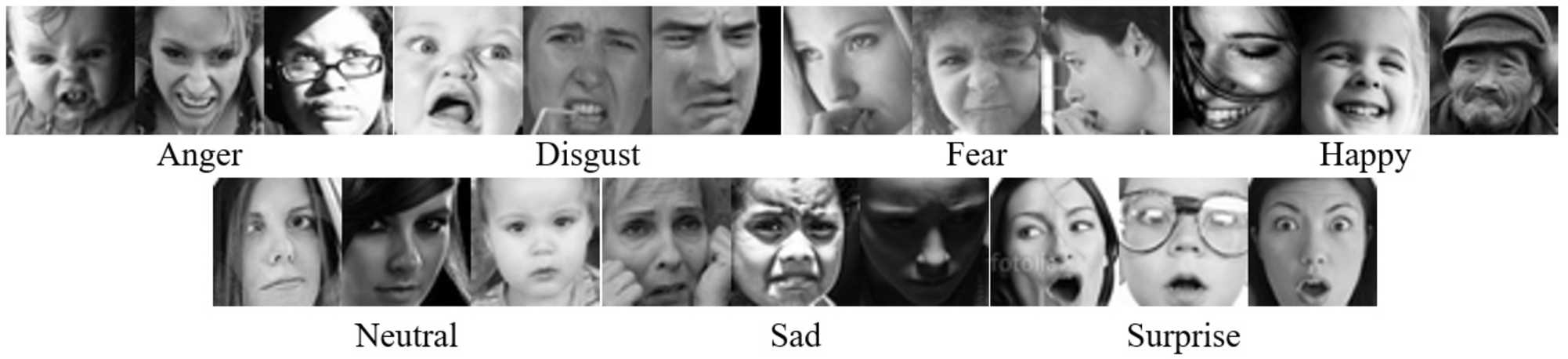}
    \caption{Part facial images with seven emotions from FER2013 dataset.}
    \label{fer2013}
\end{figure}

For real scenes, FER2013 dataset \cite{goodfellow_challenges_2013}, RAF-DB dataset \cite{li_reliable_2017} and AffectNet dataset \cite{mollahosseini_affectnet_2017} are utilized to conduct comparative experiments of different methods for FER. Specifically, FER2013 dataset \cite{goodfellow_challenges_2013} is conducted via gray images and it is shown at the International Conference on Machine Learning (ICML) in 2013. It consists of 28,709 training images and 3,589 public test images, where size of each image is $48 \times 48$. It includes scene images of lighting, occlusion and expression intensity. Its part images can be shown in Fig. \ref{fer2013}.

\begin{figure}[h]
    \centering
    \includegraphics[width=\linewidth]{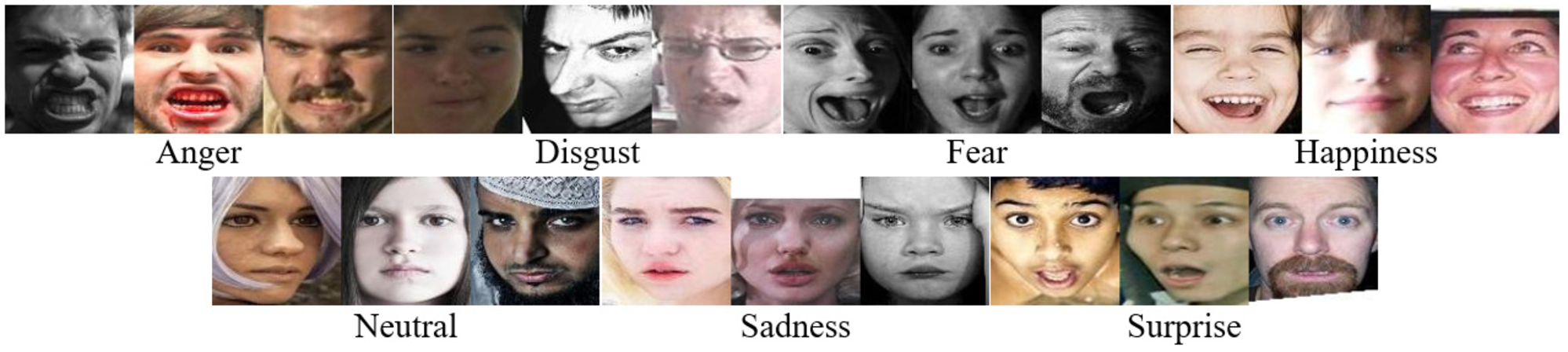}
    \caption{Part facial images with seven emotions from RAF-DB dataset.}
    \label{rafdb}
\end{figure}

RAF-DB dataset \cite{li_reliable_2017} contains approximately 30,000 face images in the real-world scenarios, where they are collected via the internet. They include seven emotions, i.e., anger, disgust, fear, happiness, neutral, sadness and surprise, where part face images can be shown in Fig. \ref{rafdb}. Specifically, this dataset was divided into 12,271 images for training and 3,068 images for validation. All images are aligned using the alignment methodology offered by RAF-DB \cite{li_reliable_2017} to ensure consistency and accuracy.

\begin{figure}[h]
    \centering
    \includegraphics[width=\linewidth]{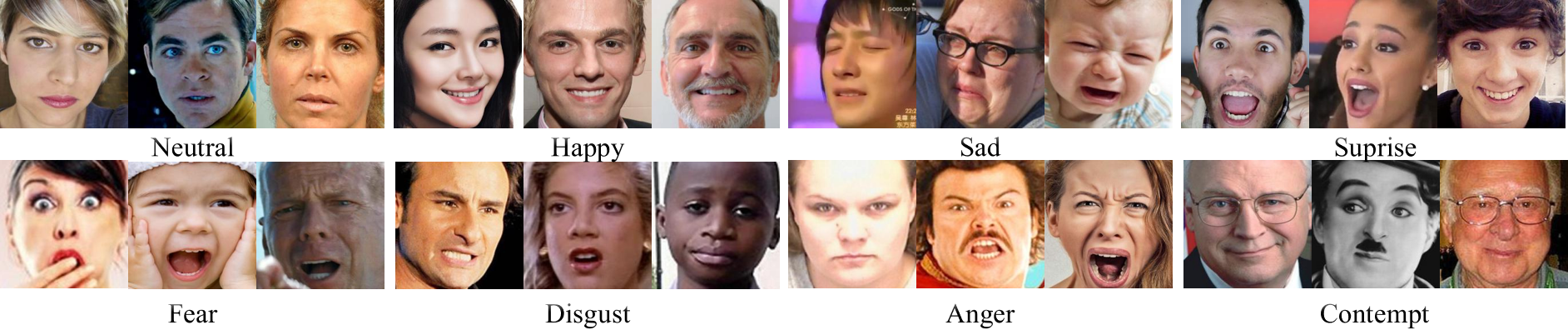}
    \caption{Part facial images with eight emotions from AffectNet dataset.}
    \label{affectnet}
\end{figure}

AffectNet dataset \cite{mollahosseini_affectnet_2017} is the largest available dataset for facial expression recognition, containing over 450,000 facial images that have been meticulously labeled by a team of expert annotators. The data was sourced using 1,250 emotion-related keywords across three major search engines. For our experiments, we conduct two subsets from AffectNet dataset. The first subset (AffectNet-7) with seven expressions includes 283,901 images for training and 3,500 images for testing. The second subset (AffectNet-8) with eight expression categories, i.e., anger, disgust, fear, happiness, neutral, sadness, surprise and contempt consists of 287,651 images for training and 4,000 images for testing. Its part visual figures can be shown Fig. \ref{affectnet}.

For context-sensitive scenes, CAER-S dataset \cite{lee_context-aware_2019} and SFEW 2.0 dataset \cite{dhall_video_2015} are used to conduct comparative experiments to test robustness of different methods for emotion recognition in a dynamic and context-sensitive environment in this paper. Specifically, CAER-S dataset was created by selecting static images from video clips in the CAER dataset \cite{lee_context-aware_2019}, which is designed for context-aware emotion recognition tasks. CAER-S contains approximately 70,000 images removing facial irrelevant images, where a final set of 48,975 images is used as a training set and 20,975 images is used as a test dataset.
\begin{figure}[h]
    \centering
    \includegraphics[width=\linewidth]{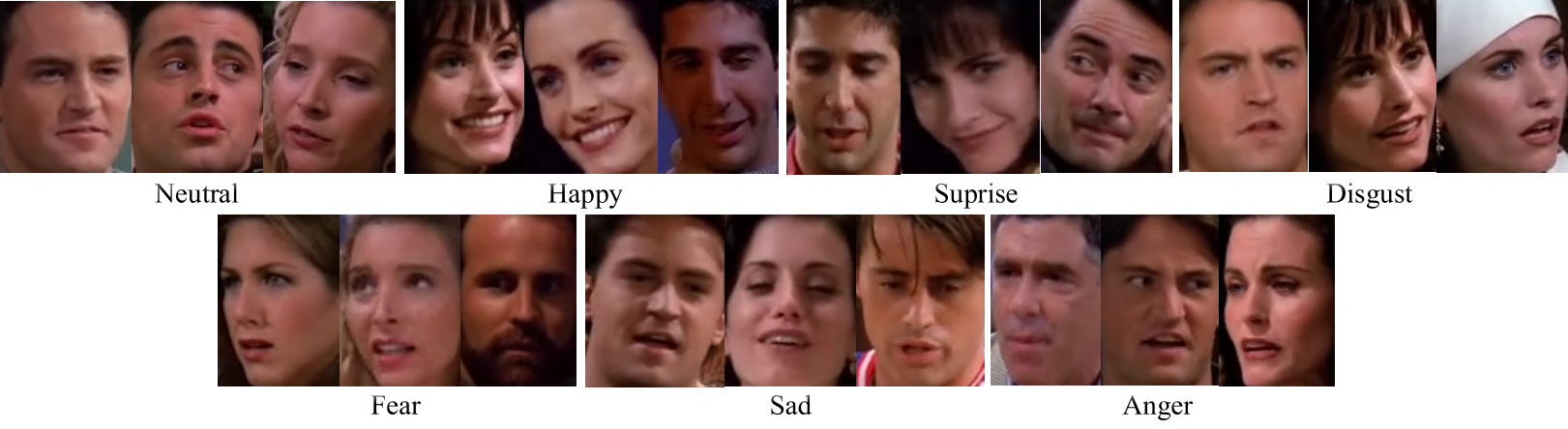}
    \caption{Part facial images with seven emotions from CAER-S dataset.}
    \label{caers}
\end{figure}

\begin{figure}[h]
    \centering
    \includegraphics[width=\linewidth]{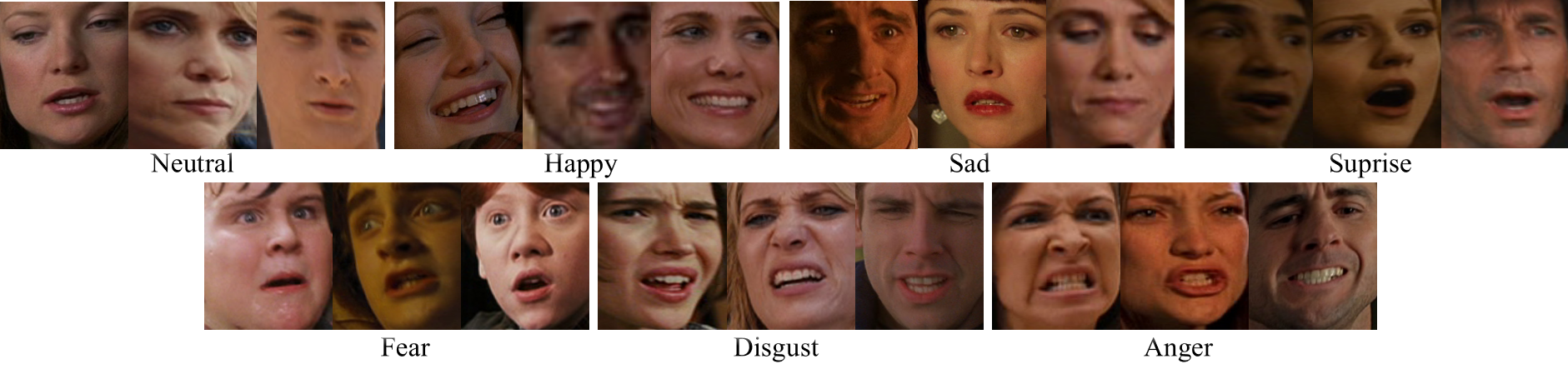}
    \caption{Part facial images with seven emotions from SFEW 2.0 dataset.}
    \label{sfew}
\end{figure}

The SFEW 2.0 dataset \cite{dhall_video_2015} was created by extracting static images from emotion-labeled video sequences in the AFEW dataset \cite{dhall_acted_2011}. This dataset contains 95 subjects with a variety of head poses, occlusions, and a broad age range to guarantee comprehensive evaluation of different FER methods under more natural and unconstrained conditions. It includes 958 images for training and 436 images for testing with seven emotions, whose visual figure can be presented via Fig. \ref{sfew}.

\subsection{Experimental settings}
All the experiments in this paper are conducted on a server equipped with an AMD EPYC 7502P 32-Core Processor, 12 G RAM, and one GPU of Nvidia GeForce GTX 3090 with Nvidia CUDA 11.7 and PyTorch 1.13.1. Our CMNet uses part parameters of pre-trained ResNet-18 model on the MS-Celeb-1M \cite{guo_ms-celeb-1m_2016} and fine-tuned our CMNet for face recognition. Also, due to difference of varying scenes, we choose different operations to conduct experiments for keeping consistency with other popular methods for FER.

For RAF-DB, FER2013, and AffectNet datasets, we use their corresponding official alignment methods \cite{li_reliable_2017,goodfellow_challenges_2013,mollahosseini_affectnet_2017} to conduct face images, respectively. For CAER-S and SFEW 2.0 datasets, we employ RetinaFace \cite{deng_retinaface_2020} to clip face images. All obtained images mentioned from six datasets were resized to a consistent resolution of $224 \times 224$. Specifically, our method on RAF-DB is optimized by an initial learning rate of 0.01, decayed by a factor of 0.1 every 15 epochs, and a batch size of 32 for training to address FER task. For AffectNet datasets, i.e., AffectNet-7 and AffectNet-8, we use the Adaptive Moment Estimation (ADAM) \cite{kingma_adam_2014}  optimizer with a learning rate of 0.0001 and a batch size of 256 to optimize our CMNet model in FER. To mitigate the challenge of class imbalance in the AffectNet dataset, we apply sampling methods, i.e., up- and down-sampling \cite{wen_distract_2023} to address underrepresented and overrepresented issues. FER2013 is only used to test performance of our CMNet for facial emotion recognition, where our CMNet is trained by using RAF-DB and AffectNet datasets. For the CAER-S dataset, our method is trained the same as Ref. \cite{zhao_learning_2021}, where we set the batch size to 128, initialize the learning rate at 0.1 and halved the learning rate every 50 epochs.  Due to SFEW 2.0 of small samples, we pre-trained our model on FER2013 before fine-tuning it on SFEW 2.0. For this fine-tuning process, we use a batch size of 16 and an initial learning rate of 0.1 and decay to 0.1 every 15 epochs. Our methods on all the datasets except AffectNet datasets are optimized via standard Stochastic Gradient Descent (SGD) \cite{robbins_stochastic_1951} with a momentum \cite{lecun_gradient-based_2002} of 0.9, a weight decay of $10^{-4}$ in this paper.

\subsection{Ablation study}
The proposed CMNet utilizes two different modals to implement a stable classifier for facial expression recognition. That is, CMNet mainly relies on three parts, i.e., a cross-modal enhancement module, salient facial information refinement module and half-face alignment optimization mechanism to obtain good performance for FER. A cross-modal enhancement module is mainly responsible for interacting two different modals of structural and biological information in FER. To prevent over-enhancement of two different modal information, a salient facial information refinement module interacts channel and spatial information in different ways to obtain multi-domain salient information to improve stability of obtained CMNet for FER. Finally, to avoid asymmetry of left and right half face images, a half-face alignment optimization mechanism is used to align left and right half face images to enhance symmetry from the cross-modal enhancement module to improve accuracy for FER. More rationality and validity analysis of key parts in the CMNet can be shown as follows.
\begin{table*}
    \centering
    \caption{Ablation study of key parts in the proposed CMNet on RAF-DB dataset.}
        \begin{tabular}{cccccccccc}
            \hline
            \multirow{2}{*}{No.} & \multirow{2}{*}{CMEM} & \multicolumn{6}{c}{SFIRM} & \multirow{2}{*}{HFAOM} & \multirow{2}{*}{Accuracy (\%)}                                                                              \\
                                 &                       & BN II                     & CBAM                   & CBAM-S4                        & CBAM-C4      & CBAM-S4C4    & CBAM-S9C9    &              &                \\ \hline
            a                    &                       &                           &                        &                                &              &              &              &              & 86.96          \\
            b                    &                       & $\checkmark$              &                        &                                &              &              &              &              & 88.05          \\
            c                    & $\checkmark$          & $\checkmark$              &                        &                                &              &              &              &              & 88.23          \\
            d                    & $\checkmark$          & $\checkmark$              &                        &                                &              &              &              & $\checkmark$ & 88.43          \\
            e                    & $\checkmark$          & $\checkmark$              & $\checkmark$           &                                &              &              &              & $\checkmark$ & 88.72          \\
            f                    & $\checkmark$          & $\checkmark$              &                        & $\checkmark$                   &              &              &              & $\checkmark$ & 88.78          \\
            g                    & $\checkmark$          & $\checkmark$              &                        &                                & $\checkmark$ &              &              & $\checkmark$ & 88.75          \\
            h                    & $\checkmark$          & $\checkmark$              &                        &                                &              & $\checkmark$ &              & $\checkmark$ & \textbf{89.11} \\
            i                    & $\checkmark$          & $\checkmark$              &                        &                                &              &              & $\checkmark$ & $\checkmark$ & 88.52          \\
            \hline
        \end{tabular}
        \label{table_ablation}
\end{table*}

\subsubsection{Cross-modal enhancement mechanism} It is known that deep networks can use different layers to mine hierarchical structural information to represent the whole face information for FER \cite{zhao_learning_2021}. Although they can obtain good performance for FER, they often overlook the intrinsic property of the face itself, which could further constrain and improve recognition accuracy. The human face possesses a symmetrical structure, not only in its morphological organization, but also in the expression itself \cite{xu_using_2013}. Taking facial symmetry into account, we use biological knowledge to guide a CNN to implement an interaction of cross-modal information containing biological information and structural information for pursuing better result of FER. Specifically, cross-modal information can be obtained via a cross-modal enhancement mechanism as well as CMEM. That is, CMEM is implemented by three parallel networks in a multi-view way to extract cross-modal information for improving accuracy of FER. The first sub-network (also regarded as Structural Block) based on the Basic Network I implemented by a residual network to act a whole face image to extract global feature. The second and third sub-networks utilize parallel networks to act left and right half face images to learn face symmetry information, where the second and third sub-networks are regarded as an upper block and lower blocks, respectively. Also, these sub-networks are implemented through Basic Network I. Interactions of three sub-networks via a concatenation operation and residual learning operation can obtain complementary information, i.e., biological information and structural information. Their effectiveness can be verified through (b) and (c) in TABLE \ref{table_ablation}.

\subsubsection{Salient facial information refinement mechanism} Due to exclusiveness of different modal information, a refinement network can be used to eliminate interference information \cite{woo_cbam_2018}. Inspired by that, a salient facial information refinement mechanism (SFIRM) is designed in this paper. SFIRM with three phases mainly uses channel and spatial information in different interaction ways to refine key information and extract salient information for FER. The first phase uses a 4-layer Basic Network II \cite{he_deep_2016} as a refinement network to refine obtained two-modal information from the CMEM, whose effectiveness can be proved by (a) and (b) in TABLE \ref{table_ablation}, where BN II refers to the Basic Network II. To extract more salient structural information, the second phase utilizes multi-domain interactions to mine more accurate network intra representative information. Firstly, we divide obtained spatial features into four parts via a central division way to construct local spatial features. Secondly, it exploits a channel attention mechanism \cite{woo_cbam_2018} to extract four local salient channel features to achieve a coarse multi-domain interaction for FER. Then, to keep the same size as the input of Basic Network II, obtained four features are jointed as a whole feature via a concatenation operation. Finally, to prevent loss of original information from Basic Network II, we use a common attention to interact refined information from Basic Network II and obtained whole feature from the last operation, where obtained features from the second phase are regarded as attention weights to guide original information from Basic Network II to further extract salient structural information. Effectiveness of spatial division mapping can be verified via comparing (e) and (f) in TABLE \ref{table_ablation}, where CBAM-S4 symbols the attention mechanism with 4-part spatial division. Also, chosen result of divided four spatial parts is listed as follows. It is known that network shape containing height and width of network input and output is fixed and regular in general, sizes, i.e., height and width of cropped blocks are required to be the same to conduct convolutional operations \cite{he_deep_2016}. Thus, cropped number of height and width in the spatial mapping are $2, 3, 4, ..., n$. Also, cropped block number of spatial features is $n^2$. When $n$ is bigger, height and width of cropped blocks are smaller, where cropped blocks have less global structural information. That results in poor performance of FER. If obtained spatial features are not divided, local structural information is lost. Its effectiveness can be presented via comparing (e) and (f) in TABLE \ref{table_ablation}. Thus, we set $n$ to 2 for cropping spatial features and block number of cropped spatial features is 4 in this paper. Its effectiveness can be verified via comparing (e) and (h) as well as (e) and (i) in TABLE \ref{table_ablation}, where CBAM-S4C4 symbols the attention mechanism with 4-part spatial division and 4-part channel division.

The third phase utilizes another interaction method to act local channel features and global features to extract more complementary information. Also, to obtain local channel features, obtained channels from the last operation utilizes a channel division mapping to divide into four parts. Then, we use a spatial attention to act four obtained channel features to learn salient information. Next, to keep the same size as the input image, we choose a concatenation operation to joint four learned spatial mapping into a whole feature mapping. Finally, we exploit a common attention mechanism to interact obtained whole spatial feature mapping and obtained whole channel feature mapping from the second phase to mine more accurate facial information for FER. Effectiveness of channel division mapping can be verified via (e) and (g) in TABLE \ref{table_ablation}, where CBAM-C4 symbols the attention mechanism with 4-part channel division.

\begin{figure}[!h]
    \centering
    \includegraphics[width=\linewidth]{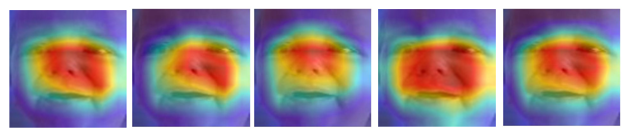}
    \caption{The attention visualization result generated by Grad CAM++ with different settings on RAF-DB dataset. The model setting from left to right is (e), (f), (g), (h), (i) in TABLE \ref{table_ablation}, respectively.}
    \label{visualization}
\end{figure}

Specifically, reason of number of divided channel features are the same as that of spatial division. In TABLE \ref{table_ablation}, we can see that (h) is higher than (e) in terms of accurate rate, which shows superiority of division number of 4 than that of non-division channel method for FER. It is more effective in terms of accuracy in comparison to division number of 9 via (h) and (i) in TABLE \ref{table_ablation}, where CBAM-S9C9 means the attention mechanism with 9-part spatial division and 9-part channel division. Global and local structural information can make a tradeoff when division number is 4. Thus, division number of 4 is optimal for channel setting in the channel division mapping. As illustrated in Fig. \ref{visualization}, the Grad CAM++ \cite{chattopadhay_grad-cam_2018} visualization results under different ablation configurations on the RAF-DB dataset demonstrate that, for this test image, all models primarily focus on the central facial region covering the eyes, nose, and mouth. However, when the proposed SFIRM is fully integrated, the high-response regions become more distinct and more tightly concentrated around these expression-relevant structures, while responses in peripheral areas are relatively suppressed. Compared with other variants, this configuration yields the most coherent and discriminative attention pattern, which indicates our proposed SFIRM is more superior for FER.

\subsubsection{Half-face alignment optimization mechanism} 
\begin{figure}[h]
    \centering
    \includegraphics[width=0.8\linewidth]{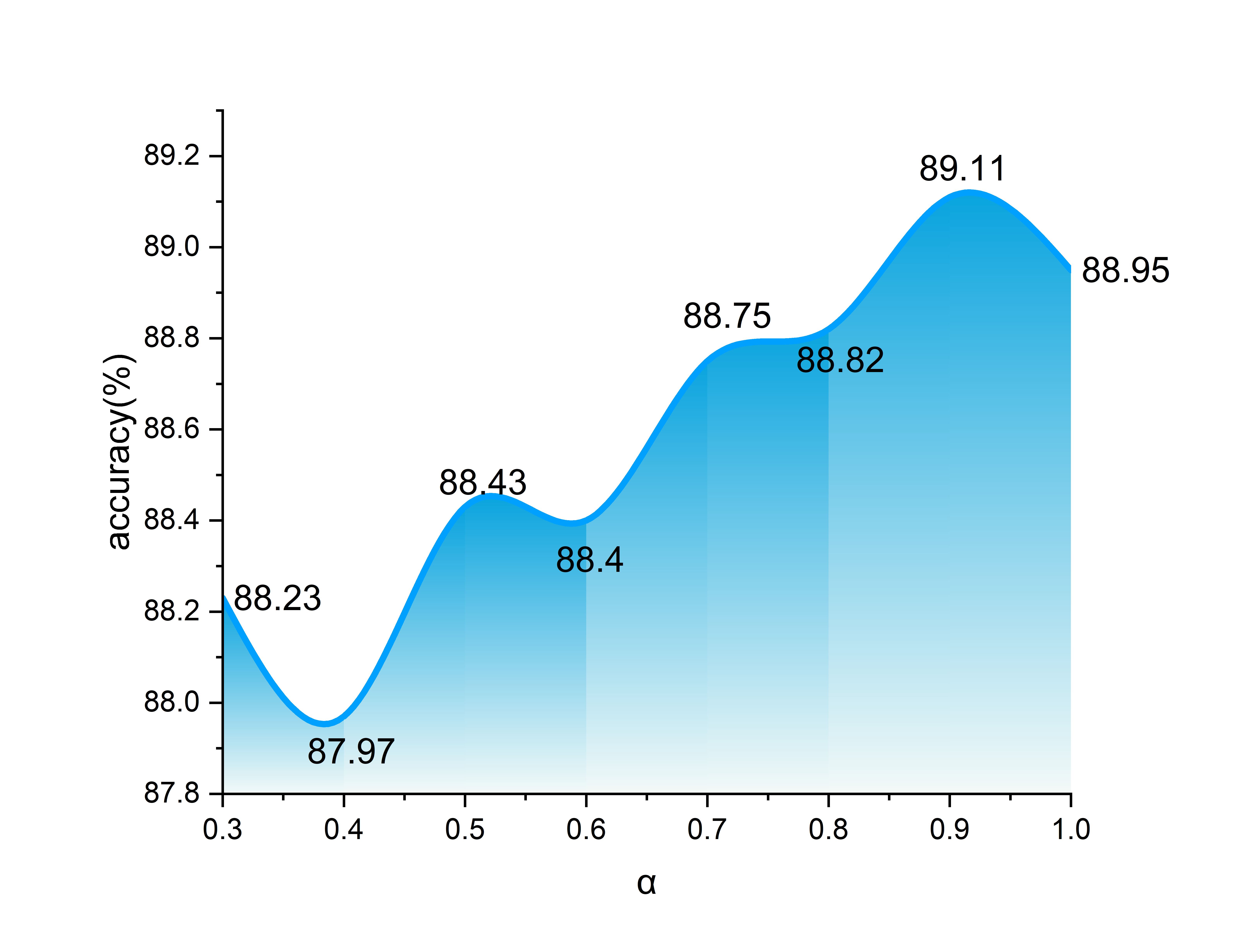}
    \caption{The accuracy on RAF-DB using different values of $\alpha$.}
    \label{alpha}
\end{figure}

To ensure symmetry of left and right half faces, a half-face alignment optimization mechanism is used to align obtained emotional information from learned left and right half faces to improve accuracy for FER. It is composed of two parts, i.e., a symmetry loss and a global loss. The symmetry loss is utilized to guide a whole face to enhance symmetry of left and right half faces for FER. The global loss is responsible for evaluating effect of our CMNet for FER. Specifically, the symmetry loss utilizes a variance function to test difference between obtained features of left and right face via UB and LB to restrict symmetry of half face images for FER. The global loss exploits outputs of the whole CMNet and given ground truth to obtain their differences to verify performance of our CMNet for FER. Two loss can use a parameter of $\alpha$ to make a trade-off between face symmetry and classifier effect of our CMNet. Detailed information of two loss functions can be shown in Section III. D. Also, effectiveness of the symmetry loss can be verified via (c) and (d) in TABLE \ref{table_ablation}. $\alpha$ is set to 0.9, which has the following reasons. Because face symmetry can assist in addressing face recognition issue, $\alpha$ is greater than 0.8 in Eq. \ref{eq4} to maintain effectiveness of FER. Also, $\alpha$ is not set to 1, which will result in invalidation of symmetry loss. Thus, $\alpha$ is set to 0.9. Its more visual effect can be shown in Fig. \ref{alpha}.

\subsection{Experimental results}
To ensure a fair comparison, we evaluated 33 publicly available methods, namely, PhaNet \cite{liu_pose-adaptive_2019}, MA-Net \cite{zhao_learning_2021}, EfficientFace \cite{zhao_robust_2021}, RAN \cite{wang_region_2020}, IPA2LT \cite{zeng_facial_2018}, SCN \cite{wang_suppressing_2020}, LAENet-SA \cite{wang_light_2021}, VGG-Net \cite{hewitt_cnn-based_2018}, AlexNet-WL \cite{mollahosseini_affectnet_2017}, FG-AGR \cite{li_fg-agr_2023}, ESR-9 \cite{siqueira_efficient_2020}, FE-SpikeFormer \cite{dong_fe-spikeformer_2025}, DENet \cite{li_unconstrained_2024}, DLN \cite{zhang_learning_2021}, gCNN \cite{li_occlusion_2018}, MMATrans+ \cite{liu_mmatrans_2024}, EAC \cite{zhang_learn_2022}, Meta-Face2Exp \cite{zeng_face2exp_2022}, KTN \cite{li_adaptively_2021}, LightExNet \cite{yang_novel_2025}, CERN \cite{li_decoding_2025}, TDF-FER \cite{najmabadi_weighted_2025}, VTFF \cite{ma_facial_2021}, ViT+SE \cite{aouayeb_learning_2021}, DMUE (ResNet-18) \cite{she_dive_2021}, PCARNet \cite{qi_novel_2024}, Baseline+BPT \cite{wang_pose-aware_2024}, CAER-Net-s \cite{lee_context-aware_2019}, ACLM \cite{she_dive_2021}, RRLA \cite{li_human_2023}, LDL-ALSG \cite{chen_label_2020}, SPWFA-SE \cite{li_facial_2020}, DAN \cite{wen_distract_2023}, ResNet-18 \cite{he_deep_2016}, ResNet-50 \cite{he_deep_2016}, MobileNet-V3-large \cite{howard_searching_2019} on five datasets: AffectNet \cite{mollahosseini_affectnet_2017}, RAF-DB \cite{li_reliable_2017}, FER2013 \cite{goodfellow_challenges_2013}, CAER-S \cite{lee_context-aware_2019} and SFEW 2.0 \cite{dhall_video_2015}. This comprehensive evaluation covers real-world and context-sensitive scenarios.

\subsubsection{Results on real-world datasets}
For real scenarios, FER2013 dataset \cite{goodfellow_challenges_2013}, RAF-DB dataset \cite{li_reliable_2017} and AffectNet dataset \cite{mollahosseini_affectnet_2017} are utilized to conduct comparative experiments of different methods for FER. Firstly, we use FER2013 dataset in a cross evaluation way to test generalization ability of our CMNet for FER. As shown in TABLE \ref{table_fer2013}, our CMNet has obtained a novel performance on AffectNet-7 and RAF-DB on FER 2013.

\begin{table}[h]
    \centering
    \caption{Accuracy of different methods on FER2013 dataset for cross evaluation}
    \begin{tabular}{ccc}
        \hline
        \multirow{2}{*}{Methods}       & \multicolumn{2}{c}{Training dataset}          \\
                                       & Affectnet-7                          & RAF-DB \\
        \hline
        SPWFA-SE \cite{li_facial_2020} & 48.68                                & 50.29  \\
        CMNet(Ours)                    & 52.10                                & 55.31  \\
        \hline
    \end{tabular}
    \label{table_fer2013}
\end{table}

\begin{figure}[!h]
    \centering
    \subfloat[]{\includegraphics[width=0.4\linewidth]{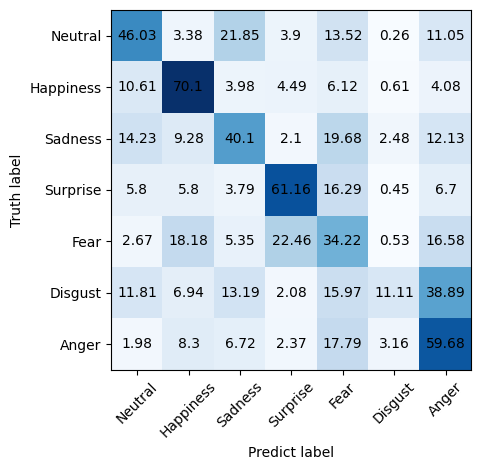}%
        \label{rafdb-fer2013}}
    \hfil
    \subfloat[]{\includegraphics[width=0.4\linewidth]{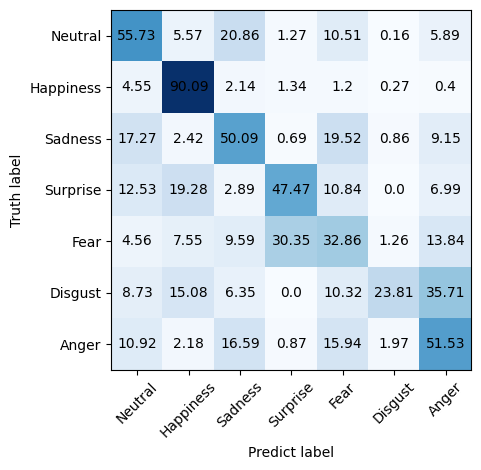}%
        \label{affectnet7-fer2013}}
    \caption{Confusion matrix of our CMNet for cross-database: (a) RAF-DB to FER2013. (b) AffetNet-7 to FER2013.}
    \label{cross-database}
\end{figure}

Fig. \ref{cross-database} presents the confusion matrix for this cross-database experiment. By analyzing the diagonal elements (excluding the neutral category), we observe that happiness, surprise, and anger achieve the highest recognition accuracy, while sadness, fear, and disgust are relatively more prone to confusion with each other. Most misclassifications occur between visually similar or semantically related categories, such as sadness with neutral, or fear/disgust with other negative emotions, which aligns with the inherent ambiguity of facial expressions under natural conditions. The distinct diagonal lines for the happiness, surprise, and anger categories indicate that our CMNet captures the characteristic eye movements and facial muscle changes associated with these expressions in a domain-invariant manner. Meanwhile, the values of the non-diagonal elements are relatively small, suggesting that the model rarely misclassifies a clearly expressed emotion into a completely different one. Furthermore, as shown in Fig. \ref{cross-database} \subref{rafdb-fer2013} and Fig. \ref{cross-database} \subref{affectnet7-fer2013}, the model exhibits a low false negative rate on most categories, highlighting its robustness under complex, real-world conditions. Combining the results from TABLE \ref{table_fer2013} with Fig. \ref{cross-database}, we can conclude that our CMNet achieves superior generalization capability for facial expression recognition in cross-database scenarios.

For RAF-DB dataset, we can see that our CMNet chooses 16 FER methods in TABLE \ref{table_rafdb} as comparative methods to test performance of our obtained classifier for FER, where our CMNet has obtained the best classification methods in facial expression recognition. That is, our CMNet improves accuracy of  0.35\% in comparison to the second method (DMUE (ResNet-18) \cite{she_dive_2021}) for FER in TABLE \ref{table_rafdb}. Our CMNet surpasses RAN \cite{wang_region_2020} by 2.21\% and MA-Net \cite{zhao_learning_2021} by 0.71\%. Also, we achieve the best performance among all methods utilizing either ResNet-18 or ResNet-50 as the backbone network.
\begin{table}[h]
    \centering
    \caption{Comparison results of different methods on the RAF-DB dataset.}
    \begin{tabular}{cc}
        \hline
        Methods                                  & Accuracy (\%)  \\
        \hline
        LightExNet \cite{yang_novel_2025}        & 85.97          \\
        CERN \cite{li_decoding_2025}             & 86.82          \\
        RAN \cite{wang_region_2020}              & 86.90          \\
        TDF-FER \cite{najmabadi_weighted_2025}   & 87.68          \\
        SCN \cite{wang_suppressing_2020}         & 88.14          \\
        VTFF \cite{ma_facial_2021}               & 88.14          \\
        ViT+SE \cite{aouayeb_learning_2021}      & 87.22          \\
        MA-Net \cite{zhao_learning_2021}         & 88.40          \\
        EfficientFace \cite{zhao_robust_2021}    & 88.36          \\
        DMUE (ResNet-18) \cite{she_dive_2021}    & 88.76          \\
        Meta-Face2Exp \cite{zeng_face2exp_2022}  & 88.54          \\
        IPA2LT \cite{zeng_facial_2018}           & 86.77          \\
        KTN \cite{li_adaptively_2021}            & 88.07          \\
        DENet \cite{li_unconstrained_2024}       & 87.35          \\
        PCARNet \cite{qi_novel_2024}             & 87.52          \\
        Baseline+BPT \cite{wang_pose-aware_2024} & 88.23          \\
        CMNet (Ours)                             & \textbf{89.11} \\
        \hline
    \end{tabular}
    \label{table_rafdb}
\end{table}

On the AffectNet dataset, we compare our method with several current FER methods in TABLE \ref{table_affectnet} on AffectNet-7 and AffectNet-8. Due to class imbalance in the training set of AffectNet containing 134,415 samples from happy and 3,750 samples from Contempt, the sampling method from \cite{wen_distract_2023} are used to address this issue. That is, an up-sampling method can overcome underrepresented categories and a down-sampling method can alleviate overrepresented categories to make a trade-off between classes in this AffectNet. As listed in TABLE \ref{table_affectnet}, we can see that our CMNet has achieved the best recognition result on AffectNet-8 and AffectNet-7 for FER. Our CMNet surpasses the current state-of-the-art (LAENet-SA \cite{wang_light_2021}) by 0.07\% in the 8-class classification. In the 7-class classification task, our CMNet outperforms the second method (MMATrans+ \cite{liu_mmatrans_2024}) by 0.31\%.  According to mentioned analysis, our method is effective for FER on real scenarios.
\begin{table}
    \centering
    \caption{Comparison results of different methods on AffectNet-8 and AffectNet-7.}
    \begin{tabular}{ccc}
        \hline
        Methods                                        & Classes & Accuracy (\%)  \\ \hline
        PhaNet \cite{liu_pose-adaptive_2019}           & 8       & 54.82          \\
        MA-Net \cite{zhao_learning_2021}               & 8       & 60.29          \\
        EfficientFace \cite{zhao_robust_2021}          & 8       & 59.89          \\
        RAN \cite{wang_region_2020}                    & 8       & 59.50          \\
        IPA2LT \cite{zeng_facial_2018}                 & 8       & 57.31          \\
        SCN \cite{wang_suppressing_2020}               & 8       & 60.23          \\
        LAENet-SA \cite{wang_light_2021}               & 8       & 61.22          \\
        VGG-Net \cite{hewitt_cnn-based_2018}           & 8       & 56.00          \\
        AlexNet-WL \cite{mollahosseini_affectnet_2017} & 8       & 58.00          \\
        FG-AGR \cite{li_fg-agr_2023}                   & 8       & 60.69          \\
        ESR-9 \cite{siqueira_efficient_2020}           & 8       & 59.30          \\
        CMNet (Ours)                                   & 8       & \textbf{61.29} \\ \hline
        FE-SpikeFormer \cite{dong_fe-spikeformer_2025} & 7       & 59.90          \\
        DENet \cite{li_unconstrained_2024}             & 7       & 60.94          \\
        DLN \cite{zhang_learning_2021}                 & 7       & 63.70          \\
        gCNN \cite{li_occlusion_2018}                  & 7       & 58.78          \\
        MA-Net \cite{zhao_learning_2021}               & 7       & 64.53          \\
        EfficientFace \cite{zhao_robust_2021}          & 7       & 63.70          \\
        MMATrans+ \cite{liu_mmatrans_2024}             & 7       & 64.89          \\
        LAENet-SA \cite{wang_light_2021}               & 7       & 64.09          \\
        EAC \cite{zhang_learn_2022}                    & 7       & 65.32          \\
        Meta-Face2Exp \cite{zeng_face2exp_2022}        & 7       & 64.23          \\
        FG-AGR \cite{li_fg-agr_2023}                   & 7       & 64.91          \\
        KTN \cite{li_adaptively_2021}                  & 7       & 63.97          \\
        CMNet (Ours)                                   & 7       & \textbf{65.63} \\ \hline
    \end{tabular}
    \label{table_affectnet}
\end{table}

\subsubsection{Results on context-sensitive datasets}
To evaluate the performance of our method in context-sensitive scenarios, we adopt the CAER-S dataset \cite{lee_context-aware_2019} and the SFEW 2.0 dataset \cite{dhall_video_2015} to conduct comparative experiments. Eight public methods in TABLE \ref{table_caers} on CAER-S dataset \cite{lee_context-aware_2019} and five public methods in TABLE \ref{table_sfew} on SFEW 2.0 dataset \cite{dhall_video_2015} are chosen as comparative methods for FER. To thoroughly assess our method, we compare it against a range of established recognition models, such as ResNet \cite{he_deep_2016} and MobileNet-V3-large \cite{howard_searching_2019}, both pre-trained on ImageNet. As illustrated in Table \ref{table_caers}, our approach attains an accuracy of 88.50\%, surpassing the deeper ResNet-50 architecture and exceeding MA-Net \cite{zhao_learning_2021} by a margin of 0.08\% on CAER-S. It is worth emphasizing that while MA-Net leverages pre-training on the large-scale MS-Celeb-1M dataset, our model is built upon a pre-trained ResNet-18 backbone. This demonstrates the efficiency and competitiveness of our method, especially under constraints of parameter size and pre-training data scale.
\begin{table}[h]
    \centering
    \caption{Comparisons of different methods on the CAER-S dataset for FER.}
    \begin{tabular}{cc}
        \hline
        Methods                                         & Accuracy (\%)  \\
        \hline
        MA-Net \cite{zhao_learning_2021}                & 88.42          \\
        EfficientFace \cite{zhao_robust_2021}           & 85.87          \\
        CAER-Net-s \cite{lee_context-aware_2019}        & 73.51          \\
        ResNet-18 \cite{he_deep_2016}                   & 86.40          \\
        ResNet-50 \cite{he_deep_2016}                   & 85.33          \\
        MobileNet-V3-large \cite{howard_searching_2019} & 85.31          \\
        ACLM \cite{she_dive_2021}                       & 87.34          \\
        RRLA \cite{li_human_2023}                       & 84.82          \\
        CMNet (Ours)                                    & \textbf{88.50} \\
        \hline
    \end{tabular}
    \label{table_caers}
\end{table}

Due to the small samples of SFEW 2.0, we first pre-trained our model on the FER2013 dataset and then fine-tuned it on SFEW 2.0 to facilitate a fair comparison. As summarized in Table \ref{table_sfew}, our method achieves a competitive accuracy of 58.26\% on this SFEW 2.0 dataset, which is the best result compared to the other methods in TABLE \ref{table_sfew}. According to mentioned analysis, our method is effective to FER on context-sensitive scenarios.
\begin{table}
    \centering
    \caption{Comparisons to different methods on the SFEW 2.0 dataset for FER.}
    \begin{tabular}{ccc}
        \hline
        Methods                             & Pretrained Dataset & Accuracy (\%)  \\ \hline
        DAN \cite{wen_distract_2023}        & MS-Celeb-1M        & 53.18          \\
        LDL-ALSG \cite{chen_label_2020}     & AffectNet, RAF     & 56.50          \\
        RAN \cite{wang_region_2020}         & MS-Celeb-1M        & 56.40          \\
        ViT+SE \cite{aouayeb_learning_2021} & FER2013            & 54.29          \\
        DENet \cite{li_unconstrained_2024}  & ImageNet           & 58.03          \\
        CMNet (Ours)                        & FER2013            & \textbf{58.26} \\
        \hline
    \end{tabular}
    \label{table_sfew}
\end{table}

\subsubsection{Complexity Analysis}
Deep learning-based methods often rely on a large number of parameters, leading to high computational overhead, evident in increased FLOPs, longer training cycles, and slower inference, which ultimately limits their practical deployment. The effectiveness and running speed are significant for the application of FER methods in real scenarios. Thus, complexity (Parameters, Running time and Flops) is essential to test our CMNet for FER.
\begin{table}[h]
    \centering
    \caption{Complexity of different methods on different databases for FER.}
    \resizebox{\linewidth}{!}{
        \begin{tabular}{ccccc}
            \hline
            Methods                                           & Parameters (M)         & Size             & Running time (ms) & FLOPs (G) \\
            \hline
            \multirow{3}{*}{CMNet(Ours)}                      & \multirow{3}{*}{11.78} & $512 \times 512$ & 67.63             & 5.83      \\
                                                              &                        & $224 \times 224$ & 14.56             & 1.12      \\
                                                              &                        & $128 \times 128$ & 9.13              & 0.36      \\
            \hline
            \multirow{3}{*}{DAN \cite{wen_distract_2023}}     & \multirow{3}{*}{19.72} & $512 \times 512$ & 44.12             & 11.67     \\
                                                              &                        & $224 \times 224$ & 9.75              & 2.23      \\
                                                              &                        & $128 \times 128$ & 8.53              & 0.73      \\
            \hline
            \multirow{3}{*}{MA-Net \cite{zhao_learning_2021}} & \multirow{3}{*}{63.54} & $512 \times 512$ & 45.45             & 10.38     \\
                                                              &                        & $224 \times 224$ & 34.19             & 3.66      \\
                                                              &                        & $128 \times 128$ & 28.75             & 1.20      \\
            \hline
        \end{tabular}
    }
    \label{table_complexity}
\end{table}

To evaluate the efficiency of CMNet, we compare its inference time, parameter count, and FLOPs with two state-of-the-art methods: DAN \cite{wen_distract_2023} and MA-Net \cite{zhao_learning_2021}. The inference time is measured by processing 32 images simultaneously at varying resolutions. As summarized in TABLE \ref{table_complexity}, CMNet achieves competitive inference speeds of 67.63 ms, 14.56 ms, and 9.13 ms for input sizes of $512 \times 512$, $224 \times 224$, and $128 \times 128$, respectively. Although DAN \cite{wen_distract_2023} attains the shortest inference times across all resolutions, our method trails by only 0.60 ms at $128 \times 128$ and 4.81 ms at $224 \times 224$. Moreover, CMNet outperforms MA-Net \cite{zhao_learning_2021} at the $128 \times 128$ and $224 \times 224$ sizes, though it is slightly slower at $512 \times 512$. Notably, as shown in TABLE \ref{table_complexity}, CMNet excels in model compactness and computational efficiency. It uses 7.94 million fewer parameters than DAN \cite{wen_distract_2023} and 51.78 million fewer than MA-Net \cite{zhao_learning_2021}. In addition, it reduces FLOPs by 1.11 G and 2.54 G compared to DAN \cite{wen_distract_2023} and MA-Net \cite{zhao_learning_2021}, respectively, at the $224 \times 224$ input size. According to mentioned illustrations, we can see that our proposed CMNet is more competitive than other popular methods, i.e., DAN \cite{wen_distract_2023} and MA-Net \cite{zhao_learning_2021} for FER on complexity, which is suitable to mobile devices.

\section{Conclusion}
In this paper, we present a cross-modal network via interacting biological and structural information to mine face symmetry information. Firstly, it uses a cross-modal enhancement mechanism to fuse face symmetry information and structural information to capture more emotional information for FER. Secondly, to overcome negative effect of obtained biological and structural information, a salient facial information refinement module interacts channel and spatial information in different ways to capture salient facial expression information to enhance robustness of the model obtained for FER. To prevent asymmetrical half-face information, a half-face alignment optimization mechanism is used to align obtained information from left and right half faces for FER. Our methods are more effective than popular methods for FER in terms of real scenes and context-sensitive scenes. In the future, we will attempt to use dynamic scenes to guide a deep network for FER.

\printbibliography

\begin{IEEEbiography}[{\includegraphics[width=1in,height=1.25in,clip,keepaspectratio]{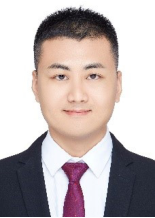}}]{Chunwei Tian} (Senior Member, IEEE) received the Ph.D. degree from Harbin Institute of Technology, Harbin, China, in 2021. He is currently a Professor with the School of Computer Science and Technology, Harbin Institute of Technology. He has published over 100 scientific papers in international journals and conferences, including IEEE Transactions on Image Processing, IEEE Transactions on Neural Networks and Learning Systems, IEEE Transactions on Multimedia, IEEE Transactions on Circuits and Systems for Video Technology, ACM Multimedia, and NeurIPS, etc. He has eight highly cited articles and five cover articles on Neural Networks and IEEE Transactions on Multimedia. His research interests include image restoration and image generation. Dr. Tian received the Distinction Prize Paper Award of Pattern Recognition. He has served as an Associate Editor for IEEE Transactions on Image Processing, IEEE Transactions on Consumer Electronics, Pattern Recognition and as an Area Editor for Computational intelligence.
\end{IEEEbiography}

\begin{IEEEbiography}[{\includegraphics[width=1in,height=1.25in,clip,keepaspectratio]{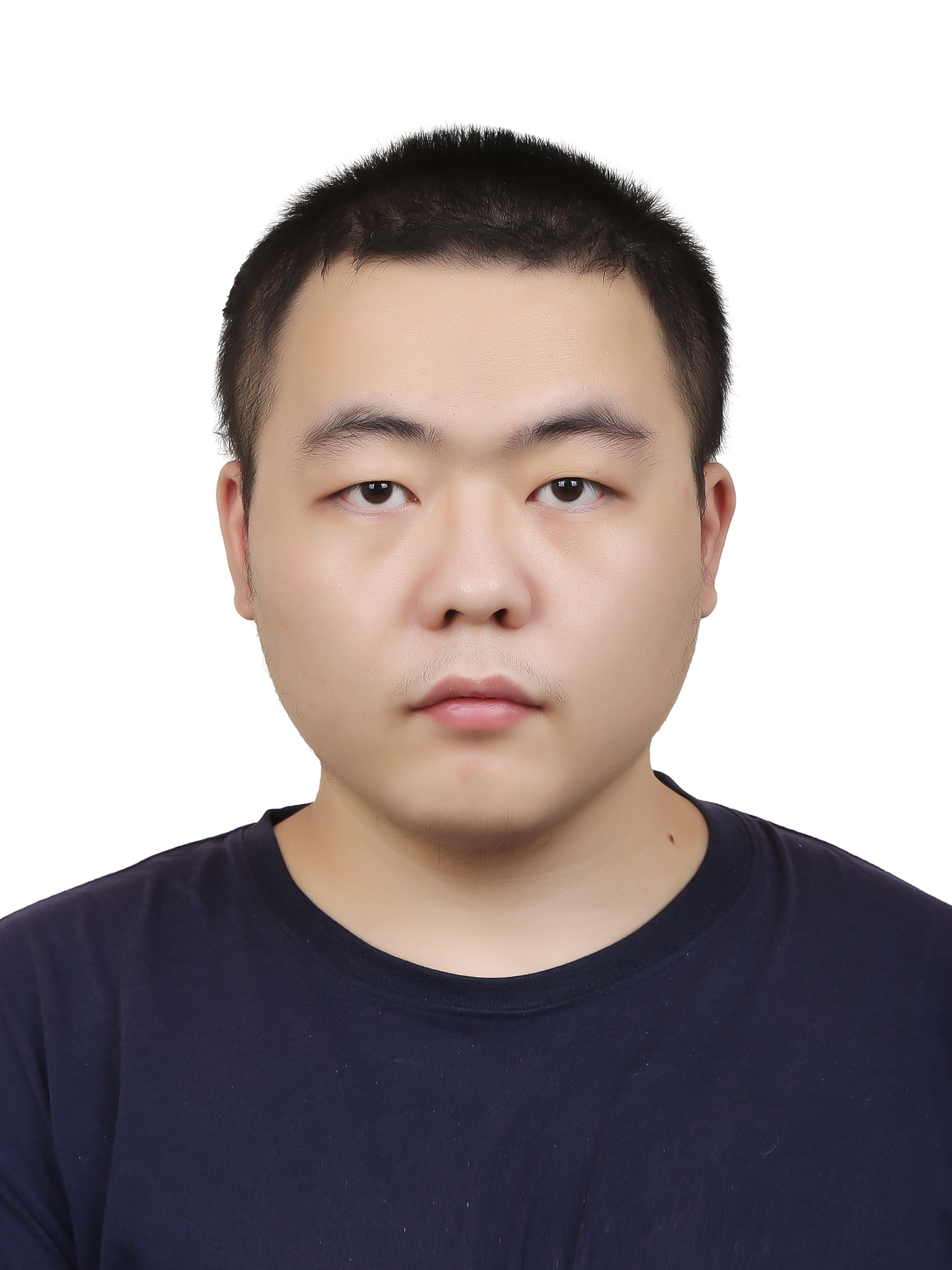}}]{Jingyuan Xie} received his B.S. degree in the School of Software, Northwestern Polytechnical University, Xian, China. He is currently working toward the M.S. degree in the School of Software, Northwestern Polytechnical University. His research interests includes image recognition and generation. He published a paper in the IEEE Transactions on Image Processing.   
\end{IEEEbiography}

\begin{IEEEbiography}[{\includegraphics[width=1in,height=1.25in,clip,keepaspectratio]{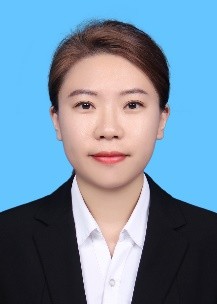}}]{Qi Zhang} received her Ph.D. degree from Harbin Institute of Technology in 2021. She is a research associate professor at School of Economics and Management, Harbin Institute of Technology at Weihai, Weihai, China. Her interests include image processing and deep learning. She has published over 20 papers containing IEEE TCE, CAAI Transactions on Intelligence Technology and Expert Systems with Application, etc. She has one ESI highly cite paper and hot paper. 
\end{IEEEbiography}

\begin{IEEEbiography}[{\includegraphics[width=1in,height=1.25in,clip,keepaspectratio]{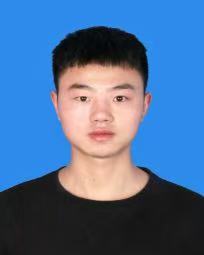}}]{Chao Li} received his M.S. degree in the School of Computer Science and Engineering, Central South University, Changsha, China.
\end{IEEEbiography}

\begin{IEEEbiography}[{\includegraphics[width=1in,height=1.25in,clip,keepaspectratio]{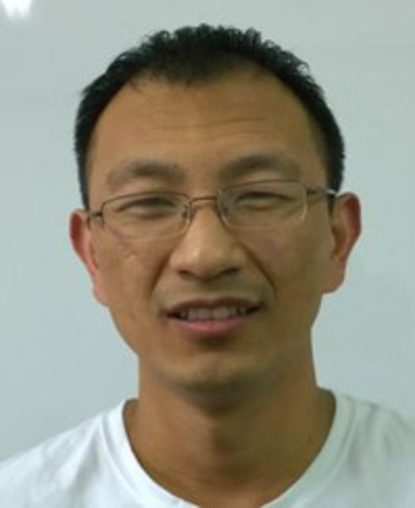}}]{Wangmeng Zuo} (Senior Member, IEEE) received the PhD degree in computer application technology from the Harbin Institute of Technology, Harbin, China, in 2007. He is currently a professor with the School of Computer Science and Technology, Harbin Institute of Technology. He has published more than 100 papers in top-tier academic journals and conferences. His current research interests include image enhancement and restoration, image and face editing, and visual generation. He has served as an associate editor for IEEE Transactions on Pattern Analysis and Machine Intelligence and IEEE Transactions on Image Processing.
\end{IEEEbiography}

\begin{IEEEbiography}[{\includegraphics[width=1in,height=1.25in,clip,keepaspectratio]{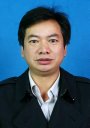}}]{Shichao Zhang} (Senior Member, IEEE) received the Ph.D. degree in computer science from Deakin University, Geelong, Australia. He is currently a China National-level-Title Professor with the Key Lab of MIMS, College of Computer Science and Technology, Guangxi Normal University, Guilin, China. His research interests include information quality and pattern discovery. He has authored or coauthored about 90 international journal papers and more than 100 international conference papers. He has won 16 national-class grants, and eight China provincial/ministerial Awards. He served/is serving as an Associate Editor for the ACM Transactions on Knowledge Discovery from Data, IEEE TRANSACTIONS ON KNOWLEDGE AND DATA ENGINEERING, Knowledge and Information Systems, and the IEEE Intelligent Informatics Bulletin, as conference Chair, PC chair, and Vice PCchair for 10 moreinternational conferences. He is a senior member of the IEEE Computer Society and a member of the ACM.
\end{IEEEbiography}

\end{document}